\pdfoutput=1
\documentclass[11pt]{article}

\usepackage[preprint]{acl}
\usepackage{times}
\usepackage{latexsym}
\usepackage[T1]{fontenc}
\usepackage[utf8]{inputenc}
\usepackage{microtype}
\usepackage{inconsolata}
\usepackage{graphicx}
\usepackage{amsmath}
\usepackage{amssymb}
\usepackage{xcolor}
\usepackage{booktabs}
\usepackage{float}
\usepackage{multirow}
\usepackage[linesnumbered,ruled,vlined]{algorithm2e}
\usepackage[capitalize]{cleveref}

\SetKwInput{KwInput}{Input}
\SetKwInput{KwOutput}{Output}
\definecolor{AlgorithmComment}{rgb}{0.35,0.35,0.70}

\title{Large-Small Model Collaboration for Enhancing Edge-Deployed Small Models}

\author{
  \textbf{Peigen Liu\textsuperscript{1}},
  \textbf{Yijiang Fan\textsuperscript{1}},
  \textbf{Zixuan Xu\textsuperscript{1}},
  \textbf{Yuren Mao\textsuperscript{1}},
  \textbf{Longbin Lai\textsuperscript{2}},
  \textbf{Ying Zhang\textsuperscript{3}} \\
  \textsuperscript{1}Zhejiang University,
  \textsuperscript{2}Tongyi Lab, Alibaba Group, \\
  \textsuperscript{3}Zhejiang Gongshang University \\
  \texttt{\{peigenliu, yijiangfan, xuzixuan, yuren.mao\}@zju.edu.cn}\\
  \texttt{longbin.lailb@alibaba-inc.com, ying.zhang@zjgsu.edu.cn}
}

\begin{document}
\maketitle

\begin{abstract}
Edge devices host domain-specific small language models (SLMs) with limited resources, while private clouds offer larger LLMs. We propose G‑Boost, an adaptive edge-cloud framework that improves a deployed SLM’s task performance without parameter updates. It formulates reasoning as a tree search, choosing at each step between SLM-only inference and SLM‑LLM logit fusion—which transfers domain knowledge from the SLM’s adapted version to the cloud LLM without exposing private data. A process reward model guides Monte Carlo tree search to select beneficial collaboration steps dynamically. The edge runs the SLM and search controller; the cloud hosts the LLM and reward model, exchanging only current context. Evaluated on GSM8K and MATH‑500 with Qwen2.5 and LLaMA2, G‑Boost outperforms the SLM alone, static fusion, and fine‑tuned baselines, gaining up to 8.6 and 10.7 percentage points over MCTS and Proxy‑Tuning, respectively. Results confirm that step‑level, reward‑guided dynamic collaboration enhances reasoning and domain utilization for deployed edge SLMs.
\end{abstract}


\section{Introduction}

\begin{figure*}[!t]
  \centering
  \includegraphics[width=\textwidth]{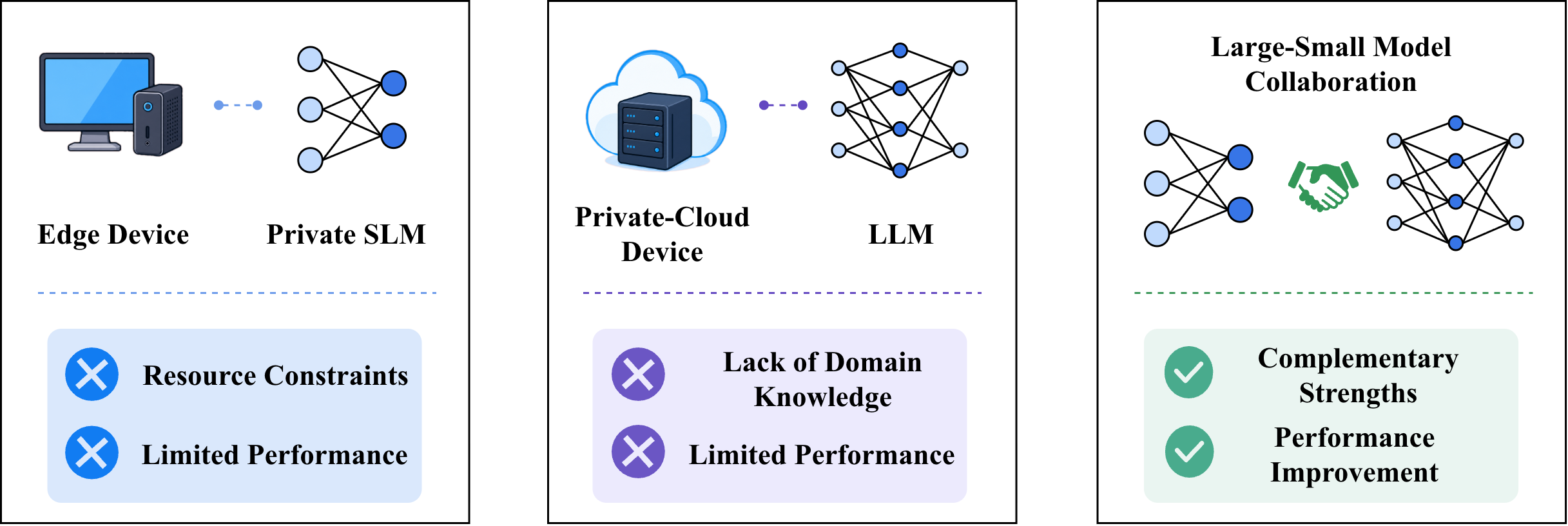}
  \par\vspace{2mm}
  \makebox[\textwidth]{%
    \makebox[0.333\textwidth]{\small (a)}%
    \makebox[0.333\textwidth]{\small (b)}%
    \makebox[0.333\textwidth]{\small (c)}%
  }
  \caption{Performance comparison among (a) an edge-deployed domain-specific small model, (b) a cloud-based large language model, and (c) large-small model collaboration.}
  \label{fig:motivation}
\end{figure*}

Edge devices are typically constrained by computing resources, storage capacity, and energy budgets and can therefore deploy only domain-specific models with relatively small parameter counts. In practical applications, users can adapt a base small language model to a domain using private domain data and deploy the trained model on a user terminal, thereby obtaining an edge-deployed domain-specific small language model (private SLM). Domain adaptation enables the model to accumulate and apply domain knowledge locally. However, edge-side resource constraints and limited model scale still restrict its problem understanding, generalization, and multi-step reasoning, thereby limiting performance on complex tasks, as shown in \cref{fig:motivation}(a). For an edge-side domain-specific small model that has already been trained and put into use, improving task-processing capability without replacing the device, retraining the model, or updating its parameters is therefore a key problem for the practical deployment of edge intelligence.

Compared with edge devices, a private cloud provides more abundant computing and storage resources and can deploy an LLM with a larger parameter count. LLMs exhibit strong capabilities in natural-language understanding and complex reasoning~\citep{GPT-4,Qwen2.5,DeepSeek-R1} and can compensate for deficiencies in an edge-side small model's problem understanding and multi-step reasoning. However, cloud-based LLMs are generally not adapted with users' private domain data. Although they possess strong general understanding and reasoning capabilities, they may lack the domain knowledge acquired by the edge-side private SLM. A cloud-based LLM therefore does not necessarily outperform an edge-side private SLM on every domain task or reasoning step; its domain-task performance is likewise constrained by this knowledge gap, as shown in \cref{fig:motivation}(b).

The edge-side private SLM and cloud-based LLM thus have clearly complementary capabilities. The former possesses strong domain knowledge but is limited by resources and model scale, whereas the latter offers stronger problem understanding and complex reasoning but lacks user-specific domain knowledge. Given this complementarity, the deployed edge-side private SLM can remain the primary task-processing model, with the cloud-based LLM invoked on demand only at steps requiring stronger problem understanding or complex reasoning. The edge model can retain and apply the user's domain knowledge, while the cloud model supplies general understanding and complex reasoning, improving task-processing performance without updating the edge model's parameters, as shown in \cref{fig:motivation}(c).

Existing studies have explored small-large model collaboration through query-level routing, token-level routing, cascaded inference, and static logit fusion~\citep{RouteLLM,ehifl,CITER,PICE,GKT,HybridLLM,AutoMix,Speculative-Decoding,allmdwss,CombLM,Proxy-Tuning,slm-llm-survey,PRISM,ConsRoute,CoMIC}, as well as multi-model routing and ensembling~\citep{RouterDC,Routoo,LLM-Blender,aeiwot,FrugalGPT}. These methods establish a foundation for exploiting complementary model capabilities. However, they often select a model once after receiving a query or apply a fixed collaboration scheme throughout generation, and they commonly imply that the cloud-based LLM is always superior. In complex tasks, different reasoning steps require different degrees of domain knowledge, language understanding, and complex reasoning. A fixed strategy cannot continuously exploit the strengths of both model classes throughout the process. A method that dynamically adjusts collaboration according to intermediate reasoning states is therefore still needed.

We argue that enhancing a deployed edge-side small model should not be viewed as a one-time model-selection problem, but should instead be modeled as a path-search problem composed of multiple reasoning actions. Based on this view, we propose G-Boost. At each reasoning step, the system can either use the edge-side private SLM to generate independently or invoke the cloud-based LLM for collaborative generation through logit fusion. A process reward model (PRM)~\citep{prm} evaluates the quality of intermediate steps, and Monte Carlo tree search (MCTS)~\citep{MCTS} uses these evaluations to explore and optimize the collaboration path. Through step-level dynamic selection, G-Boost allows the edge model to retain domain knowledge while obtaining the cloud model's understanding and reasoning capabilities on demand.

In the system implementation, the base SLM, trained private SLM, and MCTS controller are deployed on the user terminal, while the larger LLM and PRM are deployed on a private-cloud server; the components collaborate through network interfaces. For experimental validation, we use two edge-cloud model combinations based on Qwen2.5 and LLaMA2 and evaluate them on GSM8K~\citep{gsm8k} and MATH-500~\citep{prm}. G-Boost achieves the highest accuracy in every setting and outperforms both independent private-SLM search and static Proxy-Tuning collaboration. This result shows that process-reward-guided, step-level dynamic collaboration can effectively improve a deployed small model's task-processing performance without parameter updates.

The main contributions of this work are threefold. First, under deployment conditions in which edge-side computing, storage, and energy are constrained, we formalize the problem of enhancing a trained and deployed domain-specific small model without updating its parameters. Second, we propose PRM-guided G-Boost, which uses MCTS to dynamically combine edge-side private-SLM inference with SLM-LLM edge-cloud collaborative inference, jointly improving domain-knowledge utilization, problem understanding, and multi-step reasoning. Third, we validate the accuracy advantage of the method in a real edge-cloud environment consisting of a user terminal and private-cloud server, and analyze the effects of key factors including the exploration constant, step length, number of iterations, collaboration probability, and expansion strategy.

\section{Related Work}

\paragraph{Collaborative inference between SLMs and LLMs.}
Small-large model collaborative inference seeks to balance inference efficiency and model performance. Query-level routing assigns an entire query to either an SLM or an LLM according to query complexity; token-level routing sends only critical tokens to the LLM; cascaded inference uses the LLM to generate an initial draft or guiding prompt that is then refined by the SLM; and speculative decoding lets the SLM generate draft tokens that the LLM verifies in a batch during one forward pass~\citep{RouteLLM,ehifl,CITER,PICE,GKT,HybridLLM,AutoMix,Speculative-Decoding,allmdwss,slm-llm-survey}. Speculative decoding guarantees outputs identical to those of the original model without degrading generation quality. In addition, CombLM and Proxy-Tuning introduce the edge-side private SLM's domain knowledge into the cloud-based LLM through logit arithmetic, enabling black-box model adaptation~\citep{CombLM,Proxy-Tuning}. Recent studies have further considered edge-cloud deployment and collaboration objectives. PRISM routes among edge, cloud, and collaborative modes according to input sensitivity; ConsRoute performs cloud-edge-device query routing according to cross-layer response consistency; and CoMIC adopts a parameter-update-free framework that combines edge-side execution with cloud-side evaluation~\citep{PRISM,ConsRoute,CoMIC}. These studies broaden the scope of edge-cloud collaboration, but they still primarily use query-level routing, trajectory-level evaluation, or fixed collaboration processes. For domain-specific tasks, a cloud-based LLM may not possess the expertise acquired by the edge model through domain training, and different reasoning steps require different degrees of domain knowledge, language understanding, and complex reasoning. Step-level dynamic collaboration for deployed edge models therefore remains necessary.

\paragraph{Multi-LLM collaboration.}
Multi-LLM collaboration integrates the specialized capabilities of multiple LLMs to improve cross-domain or complex-task performance. Its emphasis is on capability complementarity rather than solely on reducing inference cost. RouterDC, Routoo, and expert-token routing dynamically select an appropriate LLM according to the input or generation state~\citep{RouterDC,Routoo,aeiwot}; LLM-Blender combines outputs from multiple LLMs through ranking and generative fusion~\citep{LLM-Blender}; and FrugalGPT cascades LLMs of different scales according to task complexity or resource constraints~\citep{FrugalGPT}. These methods mainly route, cascade, or ensemble multiple cloud-based LLMs, whereas this work considers step-level dynamic collaboration between a single edge-side private SLM and a cloud-based LLM. Multi-LLM collaboration generally provides limited gains on a single domain task and rarely surpasses a domain expert, making it difficult to substantially improve a domain-specific small model.

\section{Problem Setup}

We consider an edge-deployed domain-specific small language model that has already been trained and deployed. Domain adaptation uses a private training dataset $D_p$. The user adapts an open-source base SLM $\pi_s^-$ on $D_p$ to obtain the trained edge-side domain-specific model $\pi_s^+$. Both $\pi_s^-$ and $\pi_s^+$ are deployed on a resource-constrained user terminal. The small model has acquired domain knowledge and basic task-processing capability, but its limited parameter count still restricts problem understanding, generalization, and multi-step reasoning.

To improve the deployed small model's task-processing capability without updating its parameters, the user can employ a cloud-based LLM $\pi_l$ that shares the same vocabulary as $\pi_s^+$. The cloud-based LLM is deployed as a private-cloud service and provides inference through a network interface. It offers strong contextual understanding and complex reasoning but lacks the user's domain knowledge. The objective of our framework is to enable adaptive collaboration between the edge-side domain-specific SLM $\pi_s^+$ and the cloud-based LLM $\pi_l$, thereby improving the deployed small model's performance on domain-specific tasks.

In the physical deployment, the base SLM, trained private SLM, and MCTS controller reside on the user terminal; the edge-side controller maintains the search tree and selects inference actions. The LLM and PRM are deployed on a private-cloud server. Edge-side private-SLM inference generates candidate steps using only the deployed small model, whereas SLM-LLM edge-cloud collaborative inference obtains the cloud LLM's inference distribution through the network. Neither the private domain training data nor the deployed model parameters need to be uploaded to the private cloud, and the small model's parameters are not updated during collaboration. The system transmits only the current query and the reasoning context required by the protocol.

\section{Methodology}

\begin{figure*}[!t]
  \centering
  \includegraphics[width=\textwidth]{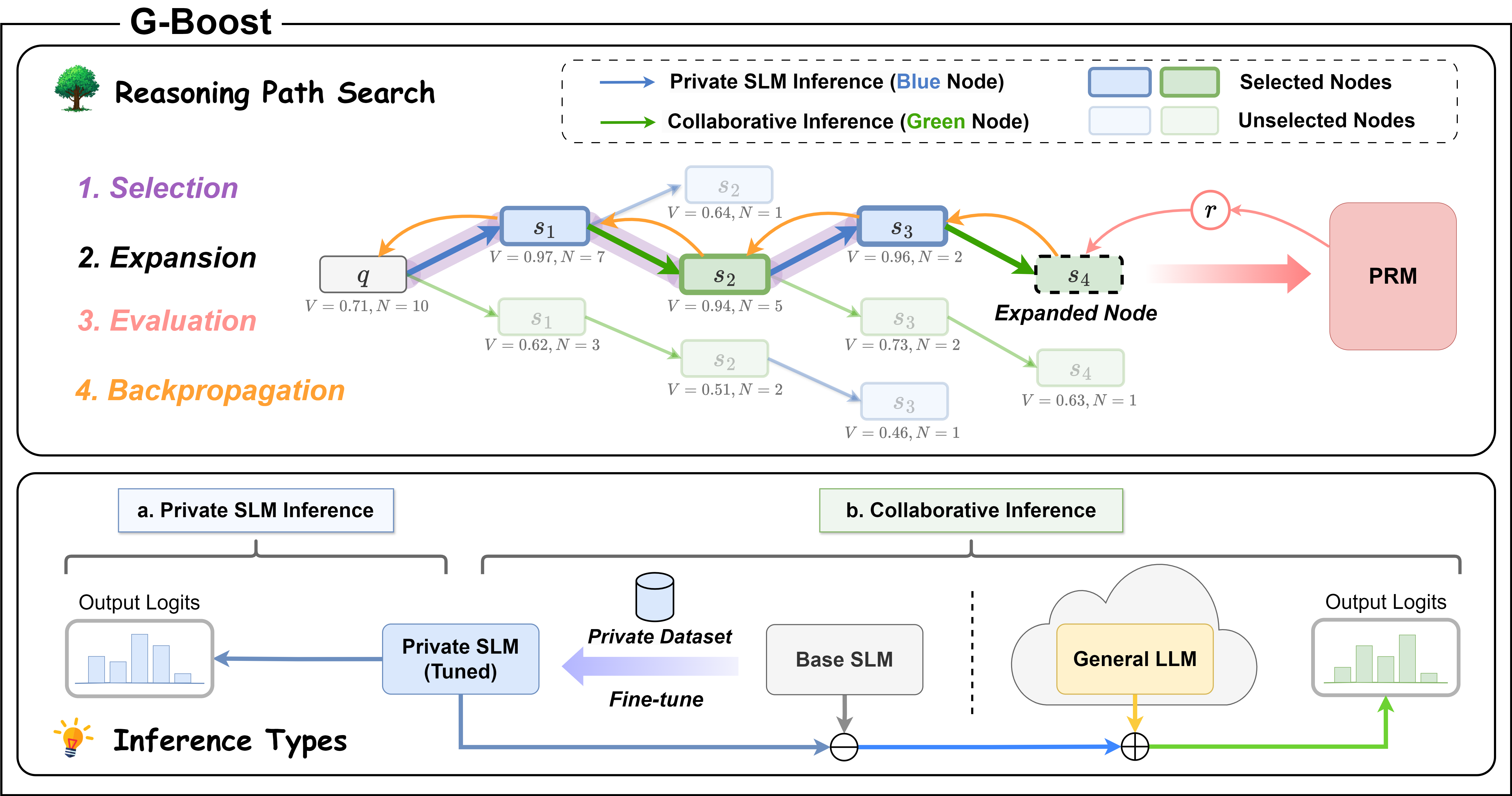}
  \caption{Overall G-Boost framework. G-Boost contains two inference modes: edge-side private-SLM inference and SLM-LLM edge-cloud collaborative inference. Guided by a process reward model, the framework uses tree search to adaptively integrate the two modes and dynamically optimize the reasoning path.}
  \label{fig:framework}
\end{figure*}

We propose G-Boost to improve the task-processing capability of an edge-deployed domain-specific small language model without updating its parameters. Its overall architecture is shown in \cref{fig:framework}. The framework dynamically integrates the domain knowledge of the edge-side private SLM with the language understanding and complex reasoning of the cloud-based LLM, enabling the small model to handle complex problems more effectively.

We model collaborative inference between the edge-side private SLM and cloud-based LLM as a search problem in a tree-structured space. The root node represents the input query, and all other nodes represent reasoning steps. We divide the reasoning process into fixed-length segments of size $L$ to define these steps. A root-to-leaf path represents the reasoning path up to a particular step. When a path reaches a final answer, its leaf is designated a terminal node, marking the end of reasoning. Each edge represents an inference action performed either independently by the edge-side private SLM or collaboratively with the cloud-based LLM. The objective is to construct the optimal reasoning path by dynamically selecting the most appropriate inference action at each step, thereby combining the advantages of the two models.

G-Boost uses MCTS to explore the collaborative reasoning space and systematically search for the optimal sequence of reasoning actions. A PRM provides fine-grained feedback on the logical consistency and task relevance of intermediate reasoning steps. During search, the PRM evaluates each newly generated step, allowing MCTS to prioritize promising paths and avoid flawed ones. Guided by process rewards, the system dynamically refines the search tree. For a query $q$, the iterative reasoning procedure constructs a search tree, with each iteration comprising selection, expansion, evaluation, and backpropagation.

\subsection{Selection}

At the beginning of each iteration, G-Boost identifies the most promising path to expand in the collaborative reasoning tree, in which the edge-side private SLM $\pi_s^+$ and cloud-based LLM $\pi_l$ collaboratively process $q$. Starting from the root, the system traverses the tree by iteratively selecting child nodes according to the Upper Confidence Bounds applied to Trees (UCT) algorithm~\citep{UCT}. This strategy balances exploiting high-quality reasoning paths against exploring underexplored ones. The UCT value of node $s$ is
\begin{equation}
\mathrm{UCT}(s)=V_s+C\sqrt{\frac{\ln N_{\operatorname{parent}(s)}}{N_s}},
\label{eq:uct}
\end{equation}
where $V_s$ is the node value accumulated during evaluation, $N_s$ is its visit count, $N_{\operatorname{parent}(s)}$ is its parent's visit count, and $C$ is the exploration constant that balances exploration and exploitation.

The trajectory from the root to the current leaf is
\begin{equation}
\tau=\{s_0,s_1,\ldots,s_d\},
\label{eq:trajectory}
\end{equation}
where $s_0$ is the root and $s_d$ is the current leaf. Among the nodes on this trajectory, candidate expansion nodes are those with unexpanded reasoning actions:
\begin{equation}
S_{\mathrm{candidate}}=\{s\mid s\in\tau\ \text{and $s$ has unexpanded actions}\}.
\label{eq:candidates}
\end{equation}
The highest-valued candidate is selected for expansion:
\begin{equation}
s_k=\underset{s\in S_{\mathrm{candidate}}}{\arg\max}\ V_s,
\label{eq:select-candidate}
\end{equation}
where $k$ denotes the index of the step in the reasoning path. By focusing on nodes with high potential value, the system prioritizes promising reasoning paths while continuing to explore less-visited paths so that potentially high-quality solutions are not missed.

\subsection{Expansion}

After selecting leaf node $s_k$, the algorithm expands the search tree by generating a new reasoning step. The newly generated child $s_{k+1}$ introduces new reasoning content and extends the tree. The two generation modes are \textbf{SLM-LLM edge-cloud collaborative inference} and \textbf{edge-side private-SLM inference}.

\subsubsection{SLM-LLM Edge-Cloud Collaborative Inference}

SLM-LLM edge-cloud collaborative inference exploits the complementary advantages of the edge-side private SLM and cloud-based LLM to improve reasoning quality. The private SLM supplies domain expertise, while the cloud-based LLM supplies stronger language understanding. This collaboration seeks to overcome the limitations of either model alone and balance generalization with specialization for more accurate reasoning.

Inspired by Proxy-Tuning~\citep{Proxy-Tuning}, we use a logit-fusion strategy to integrate the output distributions of the edge-side private SLM and cloud-based LLM. At each decoding step, the system refines the cloud-based LLM's logits by incorporating the logit offset derived from the fine-tuned edge-side private SLM and its base version. This refinement aligns the cloud model's predictions with the private SLM's domain knowledge while retaining the cloud model's robust contextual reasoning and generalization.

In this mode, the user invokes cloud-based LLM $\pi_l$. The system generates step $s_{k+1}$ by combining the logit distributions of private SLM $\pi_s^+$ and cloud-based LLM $\pi_l$. The next-token distribution is
\begin{align}
\widetilde{P}(x_t\mid q,s_{1:k},s_{k+1}^{<t})
&=\operatorname{softmax}\Big(z_c(x_t\mid q,s_{1:k},s_{k+1}^{<t}) \nonumber\\
&\quad+z_e^+(x_t\mid q,s_{1:k},s_{k+1}^{<t}) \nonumber\\
&\quad-z_e^-(x_t\mid q,s_{1:k},s_{k+1}^{<t})\Big),
\label{eq:token-fusion}
\end{align}
where $s_{k+1}^{<t}$ denotes the tokens already generated for step $s_{k+1}$ before position $t$. The probability of generating the complete step is
\begin{equation}
\widetilde{\pi}(s_{k+1}\mid q,s_{1:k})=
\prod_{t=1}^{L}\widetilde{P}(x_t\mid q,s_{1:k},s_{k+1}^{<t}).
\label{eq:step-probability}
\end{equation}
The next step is sampled from this distribution:
\begin{equation}
s_{k+1}\sim\widetilde{\pi}(\cdot\mid q,s_{1:k}),
\label{eq:collaborative-sampling}
\end{equation}
where $s_{k+1}$ is generated entirely by the collaborative probability distribution based on query $q$ and the previously generated sequence $s_{1:k}$.

\subsubsection{Edge-Side Private-SLM Inference}

Although collaborative inference is intended to exploit both models' complementary strengths, cloud-based LLM $\pi_l$ often lacks domain-specific knowledge, which may cause errors or unstable reasoning. We therefore introduce edge-side private-SLM inference as an alternative mode for generating the next step, relying solely on the fine-tuned private SLM.

In this mode, edge-side private SLM $\pi_s^+$, fine-tuned on the user's private domain dataset $D_p$, generates reasoning steps without involving the cloud-based LLM. Although constrained by its smaller scale, this mode avoids errors introduced by the cloud model and stabilizes reasoning when collaborative inference is unreliable, thereby enriching the overall search process. The next reasoning step is generated according to
\begin{equation}
s_{k+1}\sim\pi_s^+(\cdot\mid q,s_{1:k}).
\label{eq:private-sampling}
\end{equation}

\subsection{Evaluation}

The evaluation stage estimates the value of the newly expanded node to assess its potential quality in subsequent reasoning. We replace conventional rollout simulation with PRM-based evaluation to assess the expanded node more efficiently and accurately. A traditional rollout must generate a complete reasoning path, whereas a PRM can directly evaluate an intermediate step, improving evaluation stability and accuracy while reducing runtime overhead.

Because PRMs have gradually become common models for reasoning-process evaluation, we use an open-source PRM deployed in the cloud to evaluate newly expanded node $s_{k+1}$. The PRM is specifically fine-tuned to provide fine-grained feedback for each reasoning step. It takes query $q$ and reasoning-step sequence $s_{1:k+1}$ as input. The reward for $s_{k+1}$ is
\begin{equation}
r=\operatorname{PRM}(q,s_{1:k+1}).
\label{eq:reward}
\end{equation}
Compared with conventional rollout simulation, the cloud PRM directly evaluates the current intermediate step, avoiding the randomness and accumulated error of full-path simulation and providing the edge-side MCTS with fine-grained feedback for comparing candidate paths.

\subsection{Backpropagation}

After obtaining the reward, we propagate it from the newly expanded node to the root and update the statistics of every node on the path. These updates refine the search tree and make subsequent search decisions more accurate. Specifically, starting from expanded node $s_{k+1}$, reward $r$ is backpropagated along the path to the root. For a node $s$ on the path, its value is updated as
\begin{equation}
V_s=\frac{(N_s-1)V_s+r}{N_s},
\label{eq:value-update}
\end{equation}
and its visit count is updated as
\begin{equation}
N_s\leftarrow N_s+1.
\label{eq:visit-update}
\end{equation}
This process gradually focuses tree search on high-value paths and adjusts the combination of edge-side private-SLM actions and SLM-LLM edge-cloud collaborative actions in later iterations, continually improving the overall reasoning path. By guiding reasoning steps with a PRM and efficiently searching the edge-cloud collaborative inference space with MCTS, G-Boost generates logically consistent, high-quality reasoning paths and improves collaborative inference. The complete procedure is given in \cref{alg:gboost}.

\begin{algorithm}[t]
\DontPrintSemicolon
\SetAlgoLined
\SetNoFillComment
\caption{G-Boost Framework}
\label{alg:gboost}
\small
\KwInput{Query $q$, private SLM $\pi_s^+$, base SLM $\pi_s^-$, LLM $\pi_l$, PRM, step length $L$, exploration constant $C$, maximum iterations $T$, collaboration probability $p_{\mathrm{collab}}$}
\textcolor{AlgorithmComment}{\# Initialize the search tree with the root node}\;
Initialize $s_0$ with $V_{s_0}=0$ and $N_{s_0}=0$\;
\textcolor{AlgorithmComment}{\# Main MCTS loop}\;
\For{$t\leftarrow1$ \KwTo $T$}{
  \textcolor{AlgorithmComment}{\# Selection}\;
  $s\leftarrow s_0$\;
  \While{$s$ is not a leaf node}{
    $s\leftarrow\underset{u\in\operatorname{children}(s)}{\arg\max}
      \operatorname{UCT}(u)$\;
  }
  $\tau\leftarrow$ path from root to $s$\;
  $\begin{aligned}
    S_{\mathrm{candidate}}\leftarrow\{s'\mid{}&s'\in\tau\text{ and}\\
    &s'\text{ has unexpanded actions}\}
  \end{aligned}$\;
  $s_k\leftarrow\underset{s'\in S_{\mathrm{candidate}}}{\arg\max}V_{s'}$\;
  \textcolor{AlgorithmComment}{\# Expansion}\;
  \eIf{$\operatorname{rand}()<p_{\mathrm{collab}}$}{
    \textcolor{AlgorithmComment}{\# SLM-LLM edge-cloud collaborative inference}\;
    Generate $s_{k+1}$ using $\widetilde{\pi}(s_{k+1}\mid q,s_{1:k})$ as in \cref{eq:token-fusion}\;
  }{
    \textcolor{AlgorithmComment}{\# Edge-side private-SLM inference}\;
    Generate $s_{k+1}$ using $\pi_s^+(\cdot\mid q,s_{1:k})$\;
  }
  Add $s_{k+1}$ as a child of $s_k$ and set $V_{s_{k+1}}=0$, $N_{s_{k+1}}=0$\;
  \textcolor{AlgorithmComment}{\# Evaluation}\;
  $r\leftarrow\operatorname{PRM}(q,s_{1:k+1})$\;
  \textcolor{AlgorithmComment}{\# Backpropagation}\;
  $s\leftarrow s_{k+1}$\;
  \While{$s\ne s_0$}{
    $N_s\leftarrow N_s+1$\;
    $V_s\leftarrow((N_s-1)V_s+r)/N_s$\;
    $s_k\leftarrow\operatorname{parent}(s)$\;
  }
}
\KwOutput{Optimal terminal reasoning path $\tau^*$ with the highest value}
\end{algorithm}

\section{Experiments}

\subsection{Datasets}

We focus on mathematical reasoning and use two widely recognized benchmarks: GSM8K and MATH-500. GSM8K contains more than 1,000 grade-school word problems designed to test basic arithmetic and problem-solving skills, whereas MATH-500 is a subset of MATH containing 500 high-school-level problems across multiple mathematical domains. To fine-tune the SLMs for the domain-specific task, we additionally use MetaMathQA~\citep{MetaMath} as the private domain training dataset. It is a high-quality mathematical reasoning dataset constructed by augmenting the training sets of GSM8K and MATH.

\subsection{Experimental Setup}

\begin{figure}[t]
  \centering
  \includegraphics[width=\columnwidth]{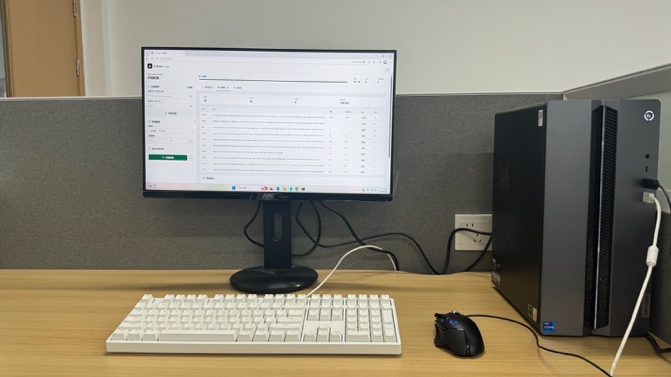}
  \par\vspace{1mm}\small (a)\par
  \includegraphics[width=\columnwidth]{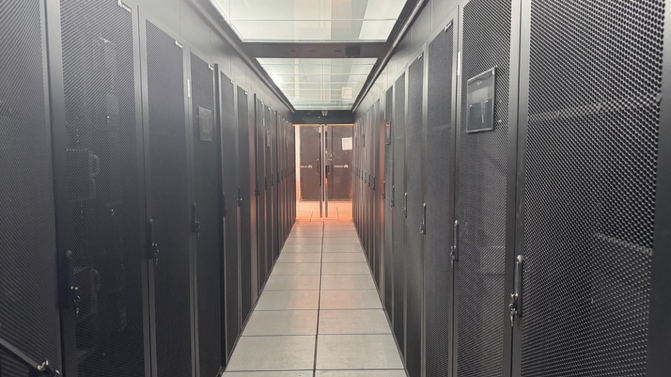}
  \par\vspace{1mm}\small (b)
  \caption{Photographs of the experimental devices: (a) the edge-side user terminal and (b) the private-cloud server.}
  \label{fig:devices}
\end{figure}

\begin{figure}[t]
  \centering
  \includegraphics[width=\columnwidth]{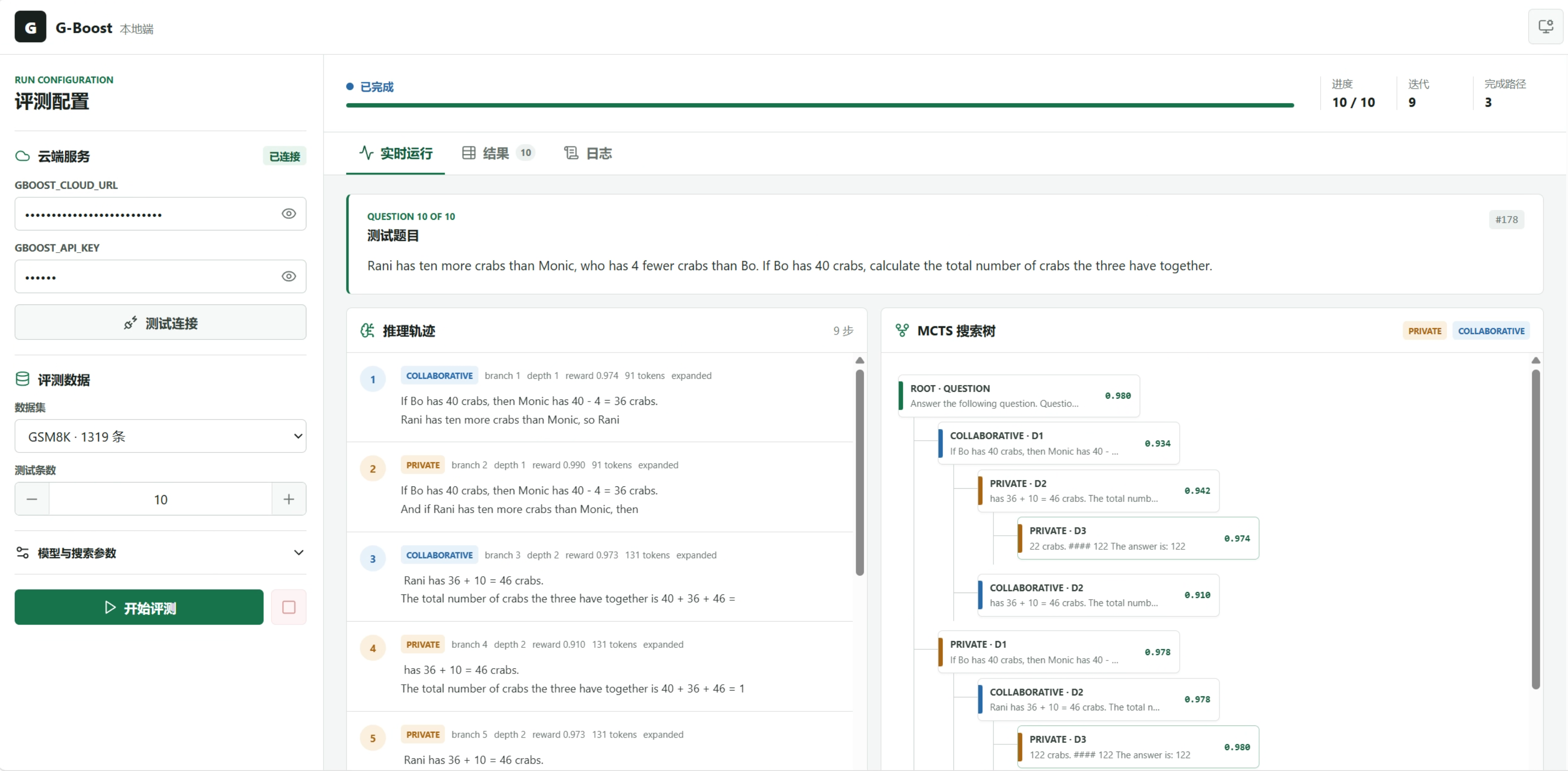}
  \par\vspace{1mm}\small (a)\par
  \includegraphics[width=\columnwidth]{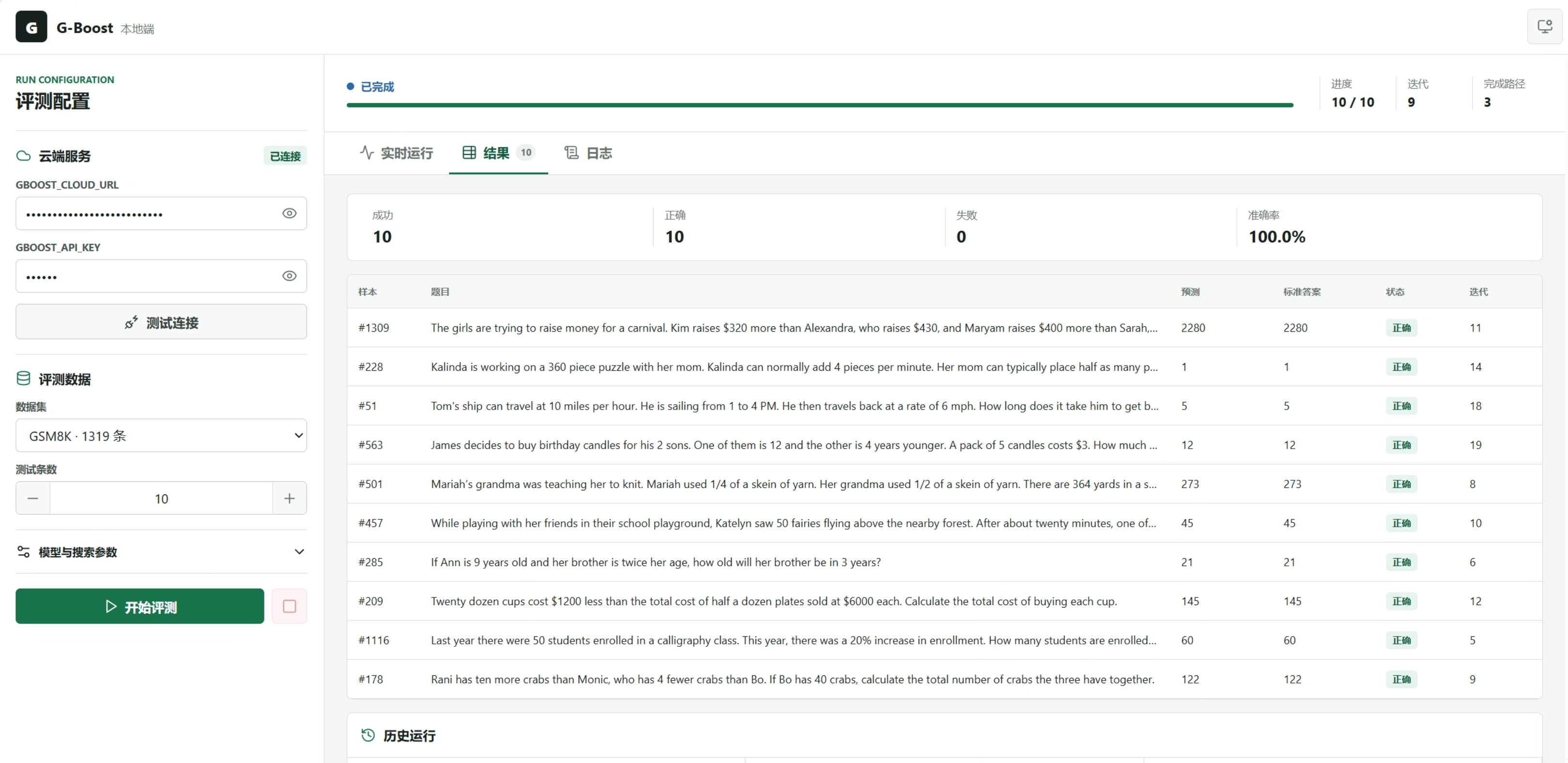}
  \par\vspace{1mm}\small (b)
  \caption{Edge-side system interface: (a) real-time reasoning trajectories and the MCTS search tree and (b) live test-problem results.}
  \label{fig:edge-ui}
\end{figure}

\begin{figure}[t]
  \centering
  \includegraphics[width=\columnwidth]{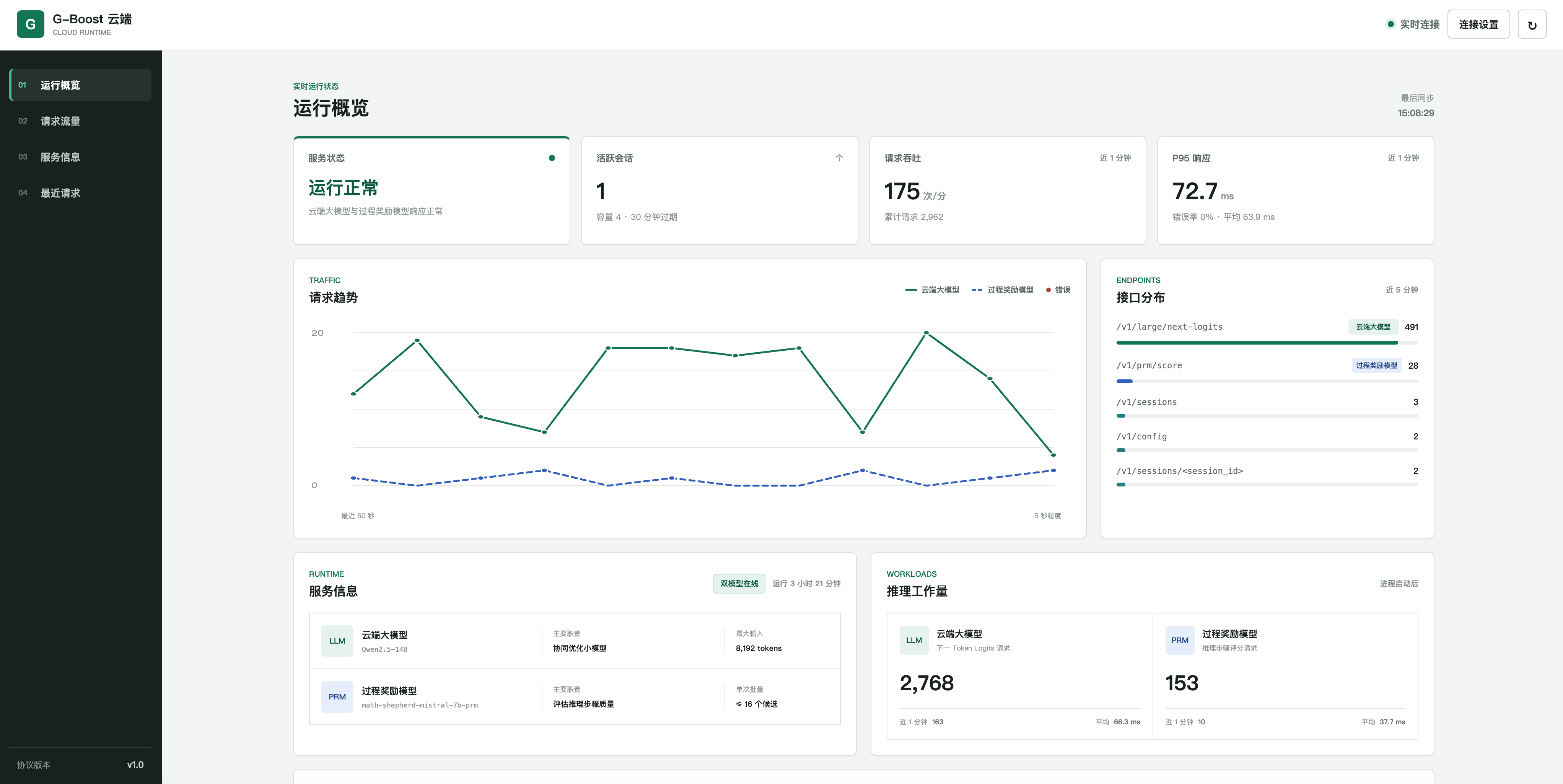}
  \caption{Private-cloud system interface.}
  \label{fig:cloud-ui}
\end{figure}

\paragraph{Edge-cloud deployment environment.}
The edge-side user terminal is a Lenovo GeekPro 17IRR with a 12th Gen Intel Core i7-12700 CPU, 12 CPU cores, 16 GB of memory, and an NVIDIA GeForce RTX 4060 Ti GPU with 8 GB of video memory. The base SLM, trained edge-side domain-specific small model, and MCTS controller are all deployed on this terminal. The private-cloud device is a PowerLeader PR4908E server equipped with two Intel Xeon Gold 6530 CPUs, providing 64 physical cores and 128 logical processors, 576 GB of memory, and six NVIDIA L40 GPUs with approximately 45 GB of video memory each. It runs 64-bit Ubuntu 24.04 LTS. The cloud-based LLM and Math-Shepherd PRM are deployed on this server and provide collaborative inference and process evaluation through network interfaces. Photographs of both devices are shown in \cref{fig:devices}; the edge-side and private-cloud system interfaces are shown in \cref{fig:edge-ui,fig:cloud-ui}. All reported accuracy results were obtained in this edge-cloud deployment environment.

\paragraph{Model settings.}
We use two edge-cloud model pairs: edge-side Qwen2.5-1.5B with cloud-based Qwen2.5-14B~\citep{Qwen2.5}, and edge-side TinyLlama-1B~\citep{TinyLlama} with cloud-based LLaMA2-13B~\citep{Llama_2}. The edge-side base SLMs are domain-tuned on MetaMathQA, while the cloud-based LLMs use their pretrained versions to reduce the potential effect of data leakage on evaluation integrity.

\paragraph{Process evaluation.}
G-Boost uses the open-source Math-Shepherd PRM~\citep{Math-Shepherd} in the cloud to evaluate intermediate mathematical reasoning steps and return reward values to the edge-side MCTS.

\paragraph{Comparison methods.}
We compare the base SLM, private SLM, fine-tuning plus MCTS guided by a PRM without edge-cloud collaboration, and Proxy-Tuning with static collaborative decoding. G-Boost and Proxy-Tuning use the same underlying logit fusion. The difference is that G-Boost uses PRM-guided MCTS to dynamically explore an action space comprising edge-side private-SLM inference and SLM-LLM edge-cloud collaborative inference. \Cref{tab:main-results} compares all methods on the two datasets.

\begin{table*}[!t]
  \centering
  \begin{tabular}{lccccccc}
    \toprule
    \multirow{2}{*}{\textbf{Model}} & \multirow{2}{*}{\textbf{Task}} & \multicolumn{3}{c}{\textbf{Edge SLM}} & \textbf{Cloud LLM} & \multicolumn{2}{c}{\textbf{Edge SLM + Cloud LLM}} \\
    \cmidrule(lr){3-5}\cmidrule(lr){6-6}\cmidrule(lr){7-8}
    & & \textbf{Base} & \textbf{Tuned} & \textbf{Tuned+MCTS} & \textbf{Base} & \textbf{Proxy-Tuning} & \textbf{G-Boost} \\
    \midrule
    \multirow{2}{*}{Qwen2.5} & GSM8K & 8.1 & 73.5 & 76.3 & 62.2 & 81.3 & \textbf{84.4} \\
    & MATH-500 & 27.2 & 33.6 & 35.8 & 31.8 & 36.8 & \textbf{44.4} \\
    \midrule
    \multirow{2}{*}{LLaMA2} & GSM8K & 1.2 & 48.2 & 59.9 & 6.5 & 54.2 & \textbf{64.9} \\
    & MATH-500 & 1.4 & 12.6 & 16.8 & 2.4 & 14.8 & \textbf{19.2} \\
    \bottomrule
  \end{tabular}
  \caption{Performance comparison on GSM8K and MATH-500. Edge SLM + Cloud LLM denotes collaborative inference methods, including Proxy-Tuning (static) and G-Boost (dynamic). Tuned+MCTS denotes an edge-side private SLM guided by MCTS without invoking the cloud-based LLM. Bold values are the best results for each task.}
  \label{tab:main-results}
\end{table*}

\subsection{Main Results}

\begin{figure}[t]
  \centering
  \includegraphics[width=\columnwidth]{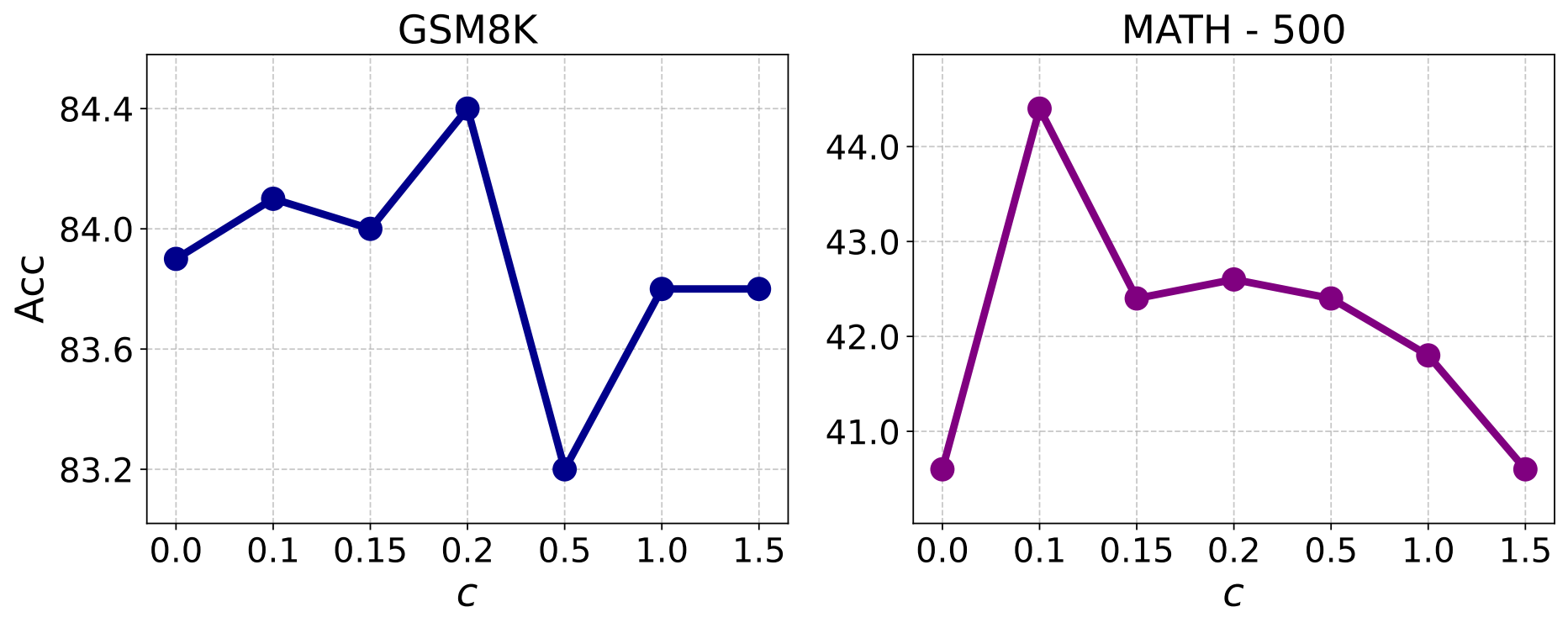}
  \caption{Effect of the exploration constant in UCT.}
  \label{fig:exploration}
\end{figure}

\begin{figure}[t]
  \centering
  \includegraphics[width=\columnwidth]{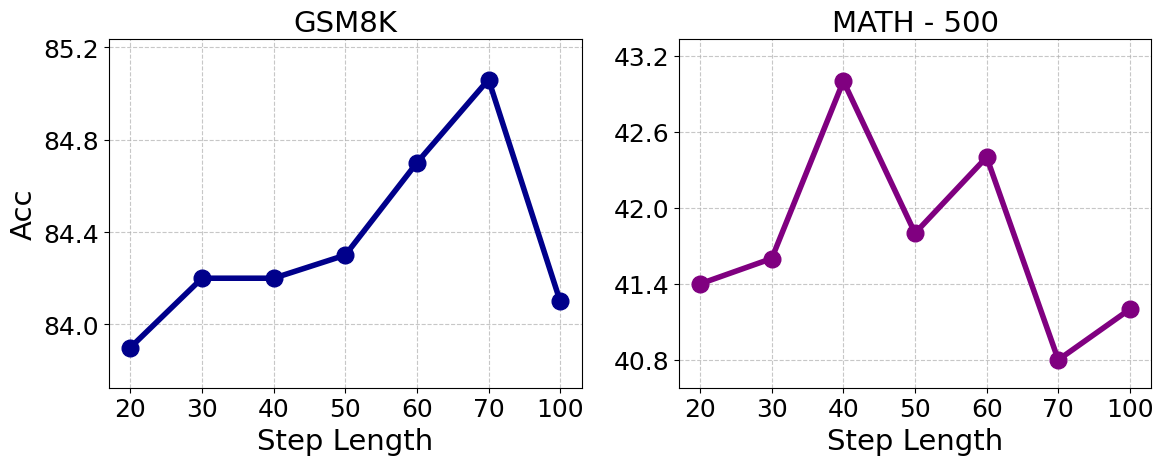}
  \caption{Effect of step length.}
  \label{fig:step-length}
\end{figure}

As shown in \cref{tab:main-results}, G-Boost achieves the highest accuracy in all four model-task settings. First, fine-tuning on private domain data substantially improves the mathematical reasoning of the edge-side base SLM, showing that domain adaptation is essential for a small model to acquire domain knowledge. Second, the cloud-based LLM does not always outperform the fine-tuned edge-side private SLM, confirming the complementarity of the two model classes. Compared with Tuned+MCTS, which relies only on the private SLM and MCTS, G-Boost improves GSM8K accuracy by 8.1 and 5.0 percentage points for Qwen2.5 and LLaMA2, respectively, and MATH-500 accuracy by 8.6 and 2.4 points. Compared with static Proxy-Tuning, G-Boost improves GSM8K accuracy by 3.1 and 10.7 points and MATH-500 accuracy by 7.6 and 4.4 points. These results show that the gain comes from process-reward-guided dynamic collaboration-path search rather than model fusion or tree search alone. The method improves domain-knowledge utilization, problem understanding, and multi-step reasoning without updating the deployed small model's parameters. The effects of the exploration constant and step length are shown in \cref{fig:exploration,fig:step-length}.

\subsection{Further Analysis}

We conduct additional analyses to better understand the behavior and internal mechanisms of G-Boost. All experiments in this subsection use the Qwen2.5 models.

\paragraph{Exploration constant in UCT.}
The exploration constant $C$ in the UCT formula balances exploration and exploitation during MCTS. We evaluate its effect on G-Boost using MATH-500 and GSM8K. As shown in \cref{fig:exploration}, performance peaks at moderate values of $C$, where the framework effectively balances exploring alternative reasoning paths and exploiting high-quality paths. Small values overemphasize exploitation and may miss optimal paths, whereas large values encourage excessive exploration and reduce efficiency. The results demonstrate the importance of tuning the exploration constant for a robust trade-off and show that the framework remains stable over a reasonable range.

\paragraph{Step length.}
The step-length parameter determines the granularity of reasoning steps in G-Boost and affects both the efficiency and quality of collaborative inference. As shown in \cref{fig:step-length}, shorter steps support more detailed reasoning but can increase the complexity of the search space, while longer steps simplify search but may overlook subtle reasoning paths. Performance remains relatively stable across a range of step lengths, indicating adaptability to different granularities. Moderate step lengths tend to perform better because they balance sufficient reasoning detail against manageable computational complexity.

\paragraph{Maximum number of iterations.}
The maximum number of iterations $T$ controls the depth of exploration in MCTS and affects the quality of the reasoning paths generated by G-Boost. As shown in \cref{tab:iterations}, increasing $T$ initially improves performance because more iterations explore the reasoning space more broadly and help identify higher-quality paths. Beyond a threshold, further increases yield only marginal gains, indicating that the framework tends to converge to stable paths within a reasonable number of iterations. This behavior is consistent with MCTS: early iterations play a greater role in shaping the search tree, whereas later iterations contribute less additional improvement. The framework is robust over a range of $T$ values and maintains relatively stable performance as $T$ varies, showing that effective reasoning does not require an excessively large iteration count.

\begin{table}[H]
  \centering
  \begin{tabular}{lccccc}
    \toprule
    $T$ & 16 & 24 & 32 & 40 & 64 \\
    \midrule
    GSM8K & 82.9 & 84.5 & 84.6 & 85.0 & \textbf{85.7} \\
    MATH-500 & 39.0 & 39.6 & \textbf{42.4} & \textbf{42.4} & 42.2 \\
    \bottomrule
  \end{tabular}
  \caption{Effect of the maximum number of iterations. Bold values are the best results in each row.}
  \label{tab:iterations}
\end{table}

\paragraph{Collaboration probability.}
The collaboration probability $p_{\mathrm{collab}}$ controls whether G-Boost invokes the cloud-based LLM during node expansion. \Cref{tab:collaboration} shows that its value substantially affects the interaction between the edge-side private SLM's domain expertise and the cloud-based LLM's stronger complex-reasoning capability. At low values, the framework relies more heavily on the private SLM, potentially limiting its ability to handle problems requiring complex reasoning or logic. At high values, the cloud-based LLM participates more often, but its lack of domain-specific knowledge can sometimes produce suboptimal results. Performance is best at moderate values, where specialization and generalization are effectively balanced. This balance allows the framework to adapt dynamically to each model's strengths, combining the private SLM's accuracy on domain tasks with the cloud-based LLM's language understanding and reasoning.

\begin{table}[H]
  \centering
  \begin{tabular}{lccccc}
    \toprule
    $p_{\mathrm{collab}}$ & 0.1 & 0.3 & 0.5 & 0.7 & 0.9 \\
    \midrule
    GSM8K & 83.7 & \textbf{84.4} & 84.1 & 84.3 & 84.2 \\
    MATH-500 & 41.8 & 42.2 & \textbf{42.4} & 41.8 & 41.2 \\
    \bottomrule
  \end{tabular}
  \caption{Effect of collaboration probability. Bold values are the best results in each row.}
  \label{tab:collaboration}
\end{table}

\paragraph{Expansion strategy.}
The expansion strategy determines how nodes are expanded during tree search and affects reasoning efficiency and quality. We compare single-child expansion, which expands one child at a time, with full-child expansion, which expands all candidate children simultaneously. As shown in \cref{tab:expansion}, single-child expansion consistently outperforms full-child expansion. Although full-child expansion can theoretically accelerate convergence by exploring multiple branches simultaneously, it may allocate the limited search budget to less promising or invalid branches, reducing attention to high-quality paths. Single-child expansion instead concentrates resources on the most promising step under PRM guidance, enabling more efficient and reliable exploration. Under a limited search budget, a more selective expansion strategy is therefore better aligned with the framework's goal of balancing domain expertise and complex reasoning.

\begin{table}[H]
  \centering
  \begin{tabular}{lcc}
    \toprule
    Expansion method & Single child & All children \\
    \midrule
    GSM8K & \textbf{84.4} & 84.3 \\
    MATH-500 & \textbf{44.4} & 42.0 \\
    \bottomrule
  \end{tabular}
  \caption{Effect of different expansion strategies. Bold values are the best results in each row.}
  \label{tab:expansion}
\end{table}

\section{Conclusion}

Under deployment conditions in which edge devices have limited computing resources, storage capacity, and energy and can host only relatively small domain-specific models, whereas a private cloud can deploy an LLM with a larger parameter count, this paper proposes G-Boost, an adaptive edge-cloud collaborative inference framework. G-Boost does not update the edge-side small model's parameters. Instead, it models collaborative task processing as a tree-structured search and uses MCTS, guided by a process reward model, to dynamically select either edge-side private-SLM inference or SLM-LLM edge-cloud collaborative inference. It thereby combines the private SLM's domain knowledge with the cloud-based LLM's language understanding and reasoning capabilities. Experiments on GSM8K and MATH-500 show that G-Boost achieves the highest accuracy in every model-task setting and outperforms the edge-side private SLM, cloud-based LLM, fine-tuning plus MCTS, and Proxy-Tuning baselines. These results verify that process-reward-guided, step-level adaptive collaboration can effectively improve the domain-knowledge utilization, problem understanding, multi-step reasoning, and overall task-processing capability of an edge-deployed domain-specific small model.

\bibliography{custom}

\end{document}